\documentclass[11pt]{article}

\usepackage[margin=1in]{geometry}
\usepackage{graphicx}
\usepackage{booktabs}
\usepackage{tabularx}
\usepackage{multirow}
\usepackage{amsmath}
\usepackage{amssymb}
\usepackage{hyperref}
\usepackage{xcolor}
\usepackage{caption}
\usepackage{subcaption}
\usepackage{enumitem}
\usepackage{placeins}
\usepackage[authoryear,round]{natbib}

\hypersetup{
  colorlinks=true,
  linkcolor=blue,
  citecolor=blue,
  urlcolor=blue
}


\title{When Are Sparse Feature Interventions Actually Localized?\\
Matched Evaluation for SAE-Based Safety Control}

\author{
  Daming Luo \quad Christy Liang \quad Junyu Xuan\\[3pt]
  University of Technology Sydney\\[3pt]
  \texttt{daming.luo@student.uts.edu.au}\\
  \texttt{Jie.Liang@uts.edu.au} \quad \texttt{Junyu.Xuan@uts.edu.au}
}

\date{}

\begin{document}
\maketitle

\begin{abstract}
Sparse autoencoder (SAE) features are increasingly proposed as localized control handles for safety-relevant behavior, on the grounds that a sparse feature intervention changes behavior more \emph{efficiently}---more behavior per unit of internal perturbation---than dense activation steering. We show that whether this advantage holds depends sharply on how the dense baseline is matched. We introduce \emph{Matched Coherence-Gated (MCG) Evaluation}, a protocol that brackets a sparse-versus-dense comparison with two complementary controls---matched target effect (fix behavior, read off perturbation) and matched perturbation norm (fix perturbation, read off behavior)---counts a jailbreak only when an output is both judge-unsafe and coherent, and cross-checks every headline number with a second behavior-completion judge. Across Gemma-2-2B/9B/27B-it and Llama-3.1-8B-Instruct with two SAE suites (Gemma Scope and Llama Scope), the protocol exposes that matching the \emph{total} perturbation norm is not sufficient: it leaves the intervention \emph{surface} unmatched, comparing a single-layer SAE ablation against an all-layer dense direction whose perturbation is diluted across the stack. On Gemma-2-9B, once we also match the surface---same-layer dense steering and dense steering projected onto the SAE decoder span---the apparent SAE efficiency advantage disappears and reverses: at every matched perturbation bin both fair baselines elicit \emph{more} coherent harmful compliance than SAE ablation (up to $-0.29$ true-jailbreak), with SAE additionally paying a capability cost at high perturbation, and the reversal holds under a HarmBench second judge. On Llama-3.1-8B and Gemma-2-27B, SAE retains a large advantage over all-layer dense steering. Independently of the dense baseline, the protocol reveals two further artifacts: high-$k$ SAE ablations trip a safety judge with incoherent text (supported by a human audit), and in a 2B model SAE ``jailbreaks'' are largely single-judge inflation that a second judge does not corroborate ($\kappa\!\to\!0$) while reasoning collapses. We conclude that SAE feature ablation is not a uniformly localized safety handle, and that sparse-localization claims should be evaluated with matched-surface, matched-basis, coherence-gated, multi-judge protocols.
\end{abstract}

\section{Introduction}

Sparse autoencoders (SAEs) provide an appealing interface for interpreting and intervening on internal representations of language models \citep{bricken2023monosemanticity,cunningham2023sparse,lieberum2024gemmascope}. If safety-relevant behavior such as refusal, harmful compliance, or jailbreak susceptibility is mediated by a small number of sparse features, then runtime feature interventions could offer a finer-grained alternative to dense activation steering or weight-level adaptation.

This promise raises a measurement problem. A safety intervention can appear successful for reasons that are not genuine localized control. A method may be weaker than the baseline it is compared against, preserving utility because it barely changes the model. A method may reach the desired target behavior only at a strength that destroys downstream capability. A high-strength intervention may also produce incoherent or degenerate text that an automated judge labels unsafe, inflating target-behavior metrics without producing meaningful harmful compliance. These failure modes matter because the core claim behind feature-level control is not merely that an intervention changes behavior, but that it changes behavior through a small and specific internal mechanism.

We therefore ask: \emph{when are SAE feature interventions actually localized?} We study this question through \emph{Matched Coherence-Gated (MCG) Evaluation}, a protocol with two complementary controls. The first matches methods by target behavior and reads off the perturbation and utility cost; the second matches methods by per-token perturbation norm and reads off the resulting behavior. Together they ask whether sparse features provide a better behavior-per-perturbation trade-off, not merely whether they can change model behavior. We define the primary target metric as unsafe \emph{and} coherent harmful compliance rather than unsafe-only judge labels, and we cross-check that metric with a second, behavior-completion judge. The main methodological contribution of this paper is therefore this matched coherence-gated evaluation protocol for activation-level safety interventions: it yields \emph{operational} evidence of localization---a measured behavior-per-perturbation advantage---rather than a claim of mechanistic causal circuit discovery.

Crucially, we run this protocol not on a single model but across a small and two large instruction-tuned models and two independent SAE suites. This turns a single-model evaluation question into a scaling question: \emph{does the localization of SAE safety control hold up as model scale and architecture change?} The results support a \emph{baseline-dependent} thesis. The apparent perturbation-efficiency of SAE ablation is highly sensitive to how the dense baseline is matched: on Gemma-2-9B it survives a naive match of total perturbation norm but reverses once the dense baseline is restricted to the same layer or to the SAE decoder span, so the headline advantage is partly an artifact of comparing a single-layer intervention against an all-layer one. On Llama-3.1-8B and Gemma-2-27B the advantage over all-layer dense steering remains large. In the small 2B model the identical recipe is not localized at all: it collapses capability and inflates a single safety judge.

The contributions of this paper are:
\begin{enumerate}[leftmargin=*]
  \item We formulate a matched coherence-gated evaluation protocol with two complementary controls---matched target-effect and matched perturbation-norm---for comparing feature-level and dense safety interventions without conflating target strength, utility loss, and degeneration artifacts.
  \item We show that matching the \emph{total} perturbation norm is not a sufficient control: it leaves the intervention \emph{surface} (the set of perturbed layers) and \emph{basis} (the subspace perturbed) unmatched. On Gemma-2-9B, against same-layer dense steering and dense steering projected onto the SAE decoder span, the apparent SAE perturbation-efficiency advantage disappears and reverses at every matched bin (true-jailbreak up to $-0.29$, with an added capability cost at high perturbation), and the reversal is confirmed by a second judge. The headline advantage under an all-layer dense baseline is therefore substantially an artifact of comparing a single-layer intervention against an all-layer one.
  \item We show that the SAE advantage over all-layer dense steering is nonetheless large on Llama-3.1-8B-Instruct and grows at Gemma-2-27B-it (both reported against an all-layer dense baseline); together with the 9B reversal, this makes baseline specification---not sparsity---the decisive factor in whether SAE looks localized.
  \item We support the coherence gate with a targeted human audit, and use a second behavior-completion judge (HarmBench) to show that high-strength SAE ablations trip a single safety judge with incoherent text, and that in Gemma-2-2B-it the SAE jailbreak signal is largely single-judge inflation ($\kappa\!\to\!0$) accompanied by reasoning collapse---a sharp negative control against extrapolating SAE steering from small models.
  \item We give feature-level diagnostics and a layer scan characterizing the refusal-aligned SAE head and its rank-decay geometry, and show that no single feature is itself a control switch.
  \item We show the efficiency comparison is \emph{direction-dependent}: for \emph{injecting} refusal as a safety improvement, dense steering is the more capability-preserving choice, so the right tool depends on the sign of the intervention rather than on sparsity alone.
\end{enumerate}

\section{Background and Related Work}

\subsection{Sparse Features as Intervention Targets}

SAEs decompose dense activations into sparse feature activations, with the goal of finding units that are more interpretable than individual neurons or arbitrary directions \citep{bricken2023monosemanticity,cunningham2023sparse}. This line of work is motivated by superposition: a model may represent more features than it has dimensions by spreading features across activation space. SAE features are attractive intervention targets because they appear to expose semantically meaningful directions that can be activated, clamped, or ablated.

Recent open SAE releases make it possible to evaluate this promise on larger language models and across architectures. Gemma Scope provides a suite of JumpReLU SAEs across Gemma 2 models and layers \citep{lieberum2024gemmascope}, and Llama Scope provides residual SAEs for Llama-3.1-8B \citep{he2024llamascope}. We use both suites, but focus on an evaluation question rather than a feature-discovery question: when a set of SAE features appears to control a safety behavior, does the effect remain localized after matching intervention strength, checking output coherence, and changing model scale or architecture?

\subsection{Activation Steering and Refusal Directions}

Dense activation steering constructs directions associated with behaviors and modifies activations along those directions at inference time. Activation engineering and representation engineering show that simple representation-level interventions can steer high-level properties such as sentiment, honesty, harmlessness, or other safety-relevant traits \citep{turner2023steering,zou2023representation}. Refusal behavior is a particularly relevant case. Prior work finds that refusal in chat models can be mediated by a low-dimensional or even single direction in residual-stream activation space \citep{arditi2024refusal}.

These dense directions are important baselines for SAE interventions. A sparse feature intervention is not convincing if it is compared only with an arbitrary dense baseline or only at an unmatched default strength. We therefore compare SAE ablations with dense refusal-direction steering under two matched controls. This comparison asks whether sparse features provide cleaner control, not merely whether they can change model behavior.

\subsection{Safety Evaluation Artifacts}

Automated safety judges are useful for scale, but they can confound harmful compliance with artifacts of generation. A response may be short, incoherent, repetitive, or otherwise degenerate while still being classified as unsafe by a judge. This problem is especially acute for interventions that directly perturb internal activations, because high-strength perturbations can damage language quality before they produce meaningful behavioral control. A single safety judge, even when paired with a coherence gate, can still systematically over- or under-count harmful compliance; a second, independently trained judge that scores behavioral \emph{completion} of the harmful request rather than surface safety provides a cross-check that is particularly important when interventions or model scale push generations off-distribution.

We use XSTest as one utility-facing benchmark because it measures exaggerated safety behavior on safe prompts \citep{rottger2023xstest}. We also report capability metrics on MMLU and GSM8K, which probe broad knowledge and multi-step mathematical reasoning \citep{hendrycks2020measuring,cobbe2021training}. For safety classification, we use a Llama Guard model as the primary automated judge \citep{fedorov2024llama}, and we cross-validate with a HarmBench behavior-completion classifier as a second judge \citep{mazeika2024harmbench}. The key methodological point is that judge labels are not treated as sufficient: the primary target metric requires both an unsafe judge label and a coherent output, and the headline conclusions are checked against a second judge.

\section{Matched Coherence-Gated Evaluation}

This paper evaluates SAE interventions through a protocol designed to separate three quantities that are often conflated: target behavior, off-target utility, and intervention-induced degeneration. The protocol is method-agnostic and can be applied to dense steering, sparse feature ablation, or other runtime activation interventions.

\paragraph{Two complementary matched controls.} Comparing a weak SAE intervention with a strong dense intervention would make SAE appear utility-preserving for the wrong reason; comparing a strong SAE intervention with a weak dense intervention would make SAE appear more effective without controlling for damage. We therefore use two matched controls that bracket the comparison from both sides:
\begin{itemize}[leftmargin=*]
  \item \textbf{Matched target-effect (matched-TE).} We sweep intervention strengths and compare methods within bins of equal coherent target behavior. This fixes \emph{behavior} and asks how much perturbation and utility loss each method spends to get there.
  \item \textbf{Matched perturbation-norm (matched-PN).} We instead bin methods by equal per-token relative residual change, and compare the resulting behavior and utility. This fixes the \emph{perturbation budget} and asks how much coherent harmful compliance each method buys per unit of internal change.
\end{itemize}
The two controls are complementary: matched-TE detects whether SAE is simply stronger or weaker, while matched-PN detects whether SAE is more \emph{perturbation-efficient}---the operational meaning of localized control. A method that ties under matched-TE but wins under matched-PN is achieving the same behavior with a smaller, more specific internal change.

\paragraph{Levels of matching.} Matched perturbation-norm is necessary but not by itself sufficient, because two interventions can share a \emph{total} per-token perturbation while differing in \emph{where} that perturbation is spent. A dense refusal direction applied across all residual layers spreads a given total norm thinly over the stack, whereas a single-layer SAE ablation concentrates the same total at one layer; at equal total norm the two are therefore not directly comparable in per-layer strength. A fair comparison must additionally match the intervention \emph{surface} (the set of layers perturbed) and, ideally, the \emph{basis} (the subspace in which the perturbation lives). We operationalize surface matching with a same-layer dense baseline that steers only at the SAE's layer, and basis matching with a dense direction projected onto the span of the top-$k$ SAE decoder vectors. Section~\ref{sec:results} shows this distinction is not academic: on Gemma-2-9B it flips the sign of the headline comparison.

\paragraph{Coherence-gated target metric.} Let \(J(x)\) indicate that the safety judge labels the response to prompt \(x\) as unsafe, and let \(C(x)\) indicate that the response is coherent. We define
\begin{equation}
\text{true\_jailbreak}(x) = \mathbf{1}\big[\,J(x)=\text{unsafe} \;\land\; C(x)=\text{coherent}\,\big].
\end{equation}
This definition does not assume that automated unsafe labels are wrong. Instead, it treats unsafe-only labels as incomplete evidence when an intervention may create degenerate text. We additionally recompute the target metric with a second behavior-completion judge to confirm that the gated signal is not an artifact of a single classifier.

\paragraph{Joint reporting.} The protocol jointly reports utility and perturbation. A localized safety intervention should not merely increase a target metric; it should do so with limited off-target capability loss and with a smaller or more specific perturbation than a dense alternative. Table~\ref{tab:protocol} summarizes the evaluation components.

\begin{table}[h]
\centering
\caption{Matched coherence-gated evaluation protocol. The protocol separates target behavior, utility retention, perturbation magnitude, and degeneration artifacts, and brackets the comparison with two complementary matched controls.}
\label{tab:protocol}
\begin{tabularx}{\linewidth}{l >{\raggedright\arraybackslash}X >{\raggedright\arraybackslash}X}
\toprule
Component & Measurement & Purpose \\
\midrule
Target effect & Unsafe-and-coherent rate (two judges) & Avoid single-judge artifacts \\
Matching (TE) & Target-effect bins & Fix behavior, compare cost \\
Matching (PN) & Perturbation-norm bins & Fix perturbation, compare behavior \\
Matching (surface/basis) & Same-layer and decoder-span dense & Match where perturbation is spent \\
Utility & MMLU, GSM8K, XSTest safe prompts & Detect off-target costs \\
Perturbation & Relative residual change per token & Quantify locality proxy \\
Uncertainty & Paired bootstrap 95\% CI (6+3 seeds) & Calibrated significance \\
Audit & Human labels on sampled outputs & Sanity-check the gate \\
\bottomrule
\end{tabularx}
\end{table}

\section{Experimental Setup}

\subsection{Models and SAEs}

Our primary target model is Gemma-2-9B-it \citep{gemma22024}, with a Gemma Scope 9B residual SAE at layer 20 (width 16k) \citep{lieberum2024gemmascope}. To test scale and architecture dependence we add two further configurations: Gemma-2-2B-it with a Gemma Scope 2B residual SAE at layer 12, and Llama-3.1-8B-Instruct \citep{dubey2024llama3} with a Llama Scope residual SAE at layer 15 \citep{he2024llamascope}. The primary safety classifier is a locally hosted Llama-Guard-3-1B model \citep{fedorov2024llama}; the second judge is a HarmBench behavior-completion classifier \citep{mazeika2024harmbench}. All interventions are applied at inference time; no model weights are updated. All model checkpoints are used under their applicable licenses and loaded from locally cached copies obtained through officially documented access paths.

\paragraph{Llama Scope loading.} The Llama Scope SAEs require care to reproduce. The release stores JumpReLU thresholds and a dataset-level activation normalization that the loader in the SAE library we used (\texttt{sae\_lens} 6.44.2) does not apply automatically. We therefore implement the forward pass manually: per-token activations are normalized to expected norm \(\sqrt{d}\), the BOS sink token (with anomalous norm \(>100\)) is excluded from normalization statistics, and the JumpReLU threshold \(0.3555\) is applied before decoding. Omitting any of these steps yields degenerate feature activations and uninterpretable interventions; we document them so the Llama-3.1-8B results are reproducible.

\subsection{Prompt Sets and Seeds}

The trigger set contains harmful or adversarial prompts designed to elicit unsafe model behavior. During prescreening, prompts are split into refused prompts and harmful-compliance prompts using the target model and the safety classifier. We use \emph{six} prompt splits with seeds 7, 13, 23, 31, 37, and 41. In each split, 800 candidate prompts are prescreened; 150 refused prompts and 108--129 harmful-compliance prompts are used to construct intervention directions, and 160 held-out refused prompts are used for safety evaluation. The refused prompts test whether an intervention removes refusal behavior; the harmful-compliance prompts provide contrast examples for constructing refusal directions. Utility is evaluated using 500 MMLU examples, 100 GSM8K examples, and 250 safe examples from XSTest per split \citep{hendrycks2020measuring,cobbe2021training,rottger2023xstest}. For the 2B and Llama-3.1-8B configurations, trigger prompts are drawn from an in-the-wild jailbreak corpus \citep{jiang2024wildjailbreak}. The primary matched perturbation-norm comparisons---Gemma-2-9B, Gemma-2-2B, and Llama-3.1-8B---use all six splits. The auxiliary and cross-scale experiments---the Gemma-2-27B scale point, the layer scan beyond layer 20, the retain audit, the single-feature ablation, the refusal-injection pairing, and the feature diagnostics---use the first three splits (seeds 7, 13, 23) unless a table caption states otherwise.

\subsection{Interventions}
\label{sec:interventions}

The dense baseline constructs a refusal direction from residual activations by subtracting the mean residual activation of harmful-compliance prompts from the mean residual activation of refused prompts. We apply this direction across residual layers and evaluate a grid of dense scaling coefficients \(\beta\). For SAE interventions, we rank SAE features within each split by cosine similarity between each decoder vector and the layer refusal direction, and ablate the top-\(k\) ranked features during inference on held-out prompts. The evaluated grid uses \(k=50\), 100, 200, 400, 800, 1200, 1600, 2400, and 3200. We also include a random SAE top1600 ablation control. Sharing a single refusal contrast source between the dense baseline and the SAE ranking is deliberate: it holds the \emph{target direction} fixed so the comparison isolates the \emph{intervention basis} (a dense direction versus a sparse feature set) rather than confounding it with a different target. We accordingly do not claim that the SAE independently discovers a novel safety concept; we test whether, given the same target, the sparse feature basis offers a better control trade-off under matched perturbation.

To test whether that advantage survives when the dense alternative is given the same intervention surface and basis, we add two stronger dense baselines, both applied only at the SAE layer. The \emph{same-layer dense} baseline steers with the refusal direction at the SAE's layer alone rather than across all layers, matching the intervention surface; here the per-token relative perturbation equals the dense coefficient $\beta$, so $\beta$ directly indexes the perturbation bin. The \emph{projected dense} baseline first projects the refusal direction onto the span of the top-$k$ SAE decoder vectors---via a QR factorization of the decoder sub-matrix---and then steers densely along the resulting direction at the SAE layer, matching the subspace available to the SAE ablation; we report the projection coverage $\lVert\mathrm{Proj}_{\text{top-}k}(r)\rVert/\lVert r\rVert$ to confirm the projected baseline is non-degenerate. Both baselines isolate whether the sparse advantage persists once the dense alternative shares the same surface and subspace.

\subsection{Metrics and Uncertainty}

We report Llama-Guard unsafe rate, coherence rate, refusal rate, retained utility, and perturbation magnitude. The primary target metric is \(\text{true\_jailbreak}\), defined in the protocol section. The automatic coherence gate uses output-length, lexical-diversity, alphabetic-ratio, and repeated-\(n\)-gram checks to flag degenerate generations. This metric treats unsafe-only labels as insufficient because a high-strength intervention can produce degenerate text that is judged unsafe without constituting meaningful harmful compliance.

For uncertainty, both safety and capability comparisons now use a paired bootstrap over per-example outcomes pooled across all six seeds for the primary comparisons (the auxiliary and cross-scale experiments use three seeds, as stated in their captions). Safety comparisons resample \((\text{split\_id}, \text{prompt\_id})\) units of the per-example \(\text{true\_jailbreak}\) outcome (960 units at six seeds). Capability comparisons resample per-example correctness on MMLU (\(\texttt{mmlu\_correct}\), 3000 units) and GSM8K (\(\texttt{gsm8k\_correct}\), 600 units), which we now store at the example level rather than as split-level aggregates. We report two-sided 95\% confidence intervals; an interval excluding zero is marked with an asterisk. The second judge is the HarmBench behavior-completion classifier, and we report the Cohen's \(\kappa\) agreement between the two judges' unsafe/yes labels on identical outputs.

\section{Results}
\label{sec:results}

\subsection{RQ1: Raw Pareto Curves Suggest Locality but Conflate Factors}

This first experiment establishes what unmatched, default-strength comparisons can and cannot show: they can suggest locality, but they cannot on their own separate target strength, perturbation size, and output quality---which motivates the matched controls in the rest of the paper. At default strengths, SAE top800 and top1600 appear cleaner than dense all-layer steering on Gemma-2-9B: they reach lower target effects with less apparent MMLU degradation. However, the high-strength SAE top3200 setting exposes the key evaluation pitfall: it has low coherence, so unsafe-only metrics would be misleading. Table~\ref{tab:raw-pareto} summarizes the raw results.\footnote{The baseline MMLU in Table~\ref{tab:raw-pareto} (0.680) is from the default-strength run reported here; the six-seed matched runs in later sections use a separate evaluation pass with baseline MMLU 0.692. The two passes differ only in sampling and seed count and are not directly comparable cell-by-cell.}

Figure~\ref{fig:target-perturb} shows that SAE feature ablation can reach comparable low-to-mid target effects with lower total relative perturbation. This result supports the possibility of localized control, but it does not by itself establish lower utility cost. A raw Pareto curve can still mix together target strength, perturbation size, and output quality. The remaining experiments therefore test whether the apparent advantage survives matched comparisons, coherence gating, a second judge, and changes of model scale and architecture.

\begin{center}
\centering
\captionof{table}{Raw 9B Pareto results. SAE top800/top1600 look favorable under default strengths, but SAE top3200 reveals a coherence-collapse failure mode.}
\label{tab:raw-pareto}
\begin{tabular}{lccc}
\toprule
Method & True jailbreak & MMLU & Coherence \\
\midrule
Baseline & 0.0000 & 0.6800 & 1.0000 \\
Dense all-layer & 0.3146 & 0.6283 & 1.0000 \\
SAE top800 & 0.1521 & 0.6900 & 1.0000 \\
SAE top1600 & 0.2187 & 0.6733 & 1.0000 \\
SAE top3200 & 0.0896 & 0.6633 & 0.3104 \\
Random SAE top1600 & 0.0146 & 0.6733 & 0.9917 \\
\bottomrule
\end{tabular}
\end{center}

\vspace{0.25em}
\begin{center}
\centering
\includegraphics[width=.90\linewidth]{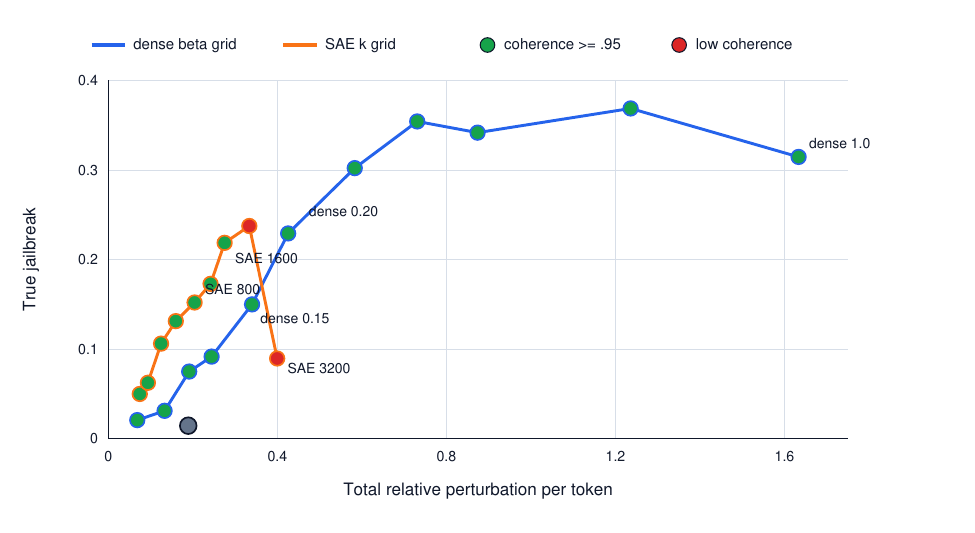}
\captionof{figure}{Target effect versus total perturbation on Gemma-2-9B. SAE reaches comparable low-to-mid target effects with lower total relative perturbation, but high-\(k\) SAE enters a coherence-collapse regime.}
\label{fig:target-perturb}
\end{center}

\subsection{RQ2: Matched-TE Ties; the Matched-PN Advantage Reverses Under Surface Matching}

\paragraph{Matched target-effect: the raw advantage ties.} To control for intervention strength, we first compare dense and SAE methods at matched target-effect points (matched-TE). Table~\ref{tab:matched-safety} reports paired bootstrap intervals for true-jailbreak differences. At all three matched-TE bins, safety differences are small and the confidence intervals cross zero. After matching by target effect, SAE and dense steering produce statistically indistinguishable rates of unsafe coherent outputs. This rules out a simple conclusion that SAE is stronger or weaker in raw target behavior.

\begin{table}[htbp]
\centering
\caption{Matched target-effect (matched-TE) safety comparisons on Gemma-2-9B. Differences are SAE minus dense, with paired bootstrap over per-example units. At equal target effect, the two methods tie.}
\label{tab:matched-safety}
\begin{tabular}{llcc}
\toprule
Matched point (TE) & Comparison & Difference & 95\% CI \\
\midrule
Target \(\approx 0.10\) & SAE top200 \(-\) dense \(\beta=0.10\) & 0.0146 & [-0.0125, 0.0417] \\
Target \(\approx 0.15\) & SAE top800 \(-\) dense \(\beta=0.15\) & 0.0021 & [-0.0292, 0.0333] \\
Target \(\approx 0.20\) & SAE top1600 \(-\) dense \(\beta=0.20\) & -0.0104 & [-0.0500, 0.0292] \\
\bottomrule
\end{tabular}
\end{table}

\paragraph{Matched perturbation-norm against all-layer dense: an apparent efficiency advantage.} The relevant question is therefore not raw strength but efficiency: at the same internal perturbation, which method produces more coherent harmful compliance? Table~\ref{tab:matched-pn-9b} bins dense and SAE by equal per-token relative residual change (matched-PN), with the dense baseline applied across all residual layers as is standard. Against this baseline the picture appears to change sharply. At low and mid perturbation, SAE produces significantly more coherent harmful compliance than dense steering---$+0.094$ and $+0.104$, both with intervals well above zero---while MMLU and GSM8K differences are indistinguishable from zero. Only at the highest perturbation bin does the advantage shrink and a capability cost appear. Read naively, this looks like the operational signature of localized control: SAE reaching the same behavior as dense with a smaller total perturbation. The next paragraph shows that this signature does not survive matching the intervention \emph{surface}.

\begin{table}[htbp]
\centering
\caption{Matched perturbation-norm (matched-PN) comparisons on Gemma-2-9B, SAE minus dense, paired bootstrap over six seeds. At equal perturbation, SAE produces significantly more coherent harmful compliance with no capability cost except at the highest bin. Units: true jailbreak 960, MMLU 3000, GSM8K 600. $^{*}$CI excludes zero.}
\label{tab:matched-pn-9b}
\resizebox{\linewidth}{!}{%
\begin{tabular}{llccc}
\toprule
Perturbation bin & Pair (SAE\,/\,dense) & True jailbreak & MMLU & GSM8K \\
\midrule
\(\approx 0.15\) & top400 / \(\beta\,0.05\)  & $+0.094^{*}$ [0.074, 0.115] & $+0.004$ [-0.005, 0.012] & $+0.010$ [-0.020, 0.040] \\
\(\approx 0.20\) & top800 / \(\beta\,0.075\) & $+0.104^{*}$ [0.083, 0.126] & $+0.000$ [-0.010, 0.011] & $-0.012$ [-0.045, 0.022] \\
\(\approx 0.30\) & top1600 / \(\beta\,0.15\) & $+0.066^{*}$ [0.039, 0.093] & $-0.034^{*}$ [-0.047, -0.020] & $-0.090^{*}$ [-0.125, -0.055] \\
\bottomrule
\end{tabular}%
}
\end{table}

\paragraph{Matching the surface and basis reverses the advantage.} The dense baseline in Table~\ref{tab:matched-pn-9b} is applied across \emph{all} residual layers, so at a matched \emph{total} perturbation it is spread thinly over the stack while the SAE ablation is concentrated at layer~20. We therefore re-run the matched perturbation-norm comparison against the two surface- and basis-matched baselines of Section~\ref{sec:interventions}, both applied at layer~20: same-layer dense steering and dense steering projected onto the top-$k$ SAE decoder span. Table~\ref{tab:matched-surface-9b} reports the result, and it reverses the headline. At all three matched bins, both fair baselines elicit \emph{more} coherent harmful compliance than SAE ablation: against same-layer dense the SAE deficit grows from $-0.034$ to $-0.286$ as perturbation increases, and against projected dense from $-0.006$ to $-0.204$, with the larger gaps excluding zero under a paired bootstrap over six seeds. At the highest bin the SAE point is additionally worse on capability (MMLU $-0.020$/$-0.028$, GSM8K $-0.122$/$-0.112$ versus same-layer/projected). The projected baseline is non-degenerate---the top-$k$ decoder span covers $0.68$/$0.75$/$0.84$ of the dense refusal direction at $k=400$/$800$/$1600$---so even restricted to the SAE's own subspace, dense steering at the matched layer is at least as effective as ablating the features. The apparent perturbation-efficiency advantage in Table~\ref{tab:matched-pn-9b} is thus largely an artifact of comparing a single-layer intervention against an all-layer one; once the dense baseline is given the same surface, the advantage on Gemma-2-9B does not survive, and the reversal holds under the HarmBench second judge.

\begin{table}[htbp]
\centering
\caption{Surface- and basis-matched comparison on Gemma-2-9B, six seeds. All methods are applied at layer~20 and binned by per-token relative perturbation. Both fair dense baselines---same-layer, and dense projected onto the top-$k$ decoder span---elicit more coherent harmful compliance than SAE ablation at every bin, under both the Llama-Guard primary judge (Guard) and the HarmBench second judge (HB). $^{*}$paired-bootstrap 95\% CI on the SAE$-$baseline true-jailbreak difference excludes zero.}
\label{tab:matched-surface-9b}
\begin{tabular}{llccc}
\toprule
Bin (rel-norm) & Method & Guard tj & HB tj & Coh. \\
\midrule
\multirow{3}{*}{$0.16$}
 & SAE top400                      & 0.135 & 0.137 & 1.00 \\
 & same-layer dense $\beta\,0.16$  & 0.170 & 0.164 & 1.00 \\
 & projected dense $k400$          & 0.142 & 0.140 & 1.00 \\
\midrule
\multirow{3}{*}{$0.20$}
 & SAE top800                      & 0.163 & 0.149 & 1.00 \\
 & same-layer dense $\beta\,0.20$  & 0.221 & 0.203 & 1.00 \\
 & projected dense $k800$          & 0.200 & 0.157 & 1.00 \\
\midrule
\multirow{3}{*}{$0.31$}
 & SAE top1600                     & 0.209 & 0.145 & 1.00 \\
 & same-layer dense $\beta\,0.31$  & 0.496 & 0.382 & 1.00 \\
 & projected dense $k1600$         & 0.414 & 0.291 & 1.00 \\
\midrule
\multicolumn{5}{l}{\emph{SAE $-$ baseline (Guard true-jailbreak, paired bootstrap, bins 0.16/0.20/0.31):}} \\
 & \,vs same-layer dense & \multicolumn{3}{l}{$-0.034^{*}$ \;/\; $-0.058^{*}$ \;/\; $-0.286^{*}$} \\
 & \,vs projected dense  & \multicolumn{3}{l}{$-0.006$ \;/\; $-0.037^{*}$ \;/\; $-0.204^{*}$} \\
\bottomrule
\end{tabular}
\end{table}

\FloatBarrier

\subsection{RQ3: The Clean Regime Is Scale- and Architecture-Dependent}

We next ask whether the advantage over all-layer dense steering---before the surface correction of Table~\ref{tab:matched-surface-9b}---is specific to one model, one layer, or one SAE suite, and whether the capability picture changes with scale. The advantage over all-layer dense persists across layers and architectures and grows at 27B, but the same recipe fails sharply for the small 2B model; we flag throughout that the large-model comparisons in this subsection remain against an all-layer baseline.

\begin{figure}[htbp]
\centering
\includegraphics[width=\linewidth]{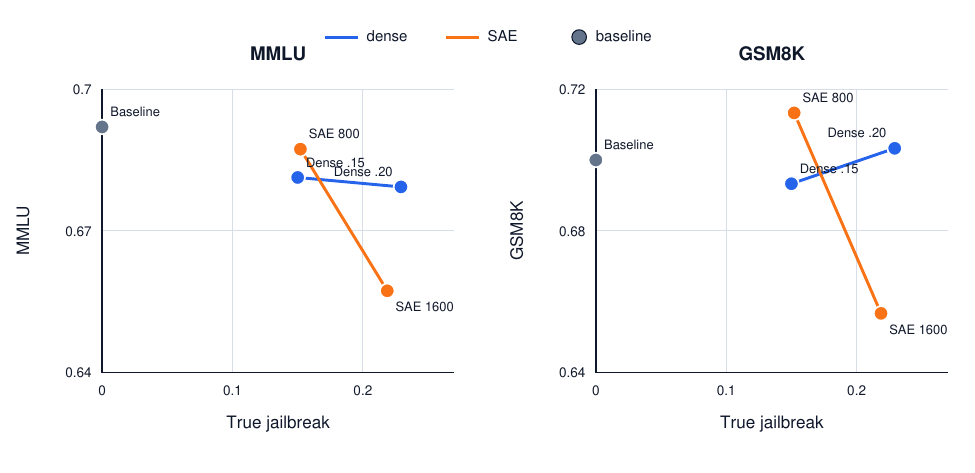}
\caption{Utility at matched target-effect points on Gemma-2-9B. SAE top800 is competitive near the lower target-effect point, but SAE top1600 loses utility relative to dense \(\beta=0.20\).}
\label{fig:matched-utility}
\end{figure}

\paragraph{Utility at matched target effect (9B).} Figure~\ref{fig:matched-utility} compares retained utility at matched target-effect points. SAE top800 is competitive with dense \(\beta=0.15\): MMLU is 0.006 higher and GSM8K is 0.020 higher. At the higher target-effect point, SAE top1600 is worse than dense \(\beta=0.20\), with MMLU lower by 0.022 and GSM8K lower by 0.0467. The six-seed matched-PN intervals in Table~\ref{tab:matched-pn-9b} sharpen this: the top1600 capability cost (MMLU $-0.034$, GSM8K $-0.090$) is now statistically significant rather than a diagnostic trend, so the regime boundary at high \(k\) is real.

\paragraph{The clean regime holds across layers.} The medium-\(k\) result is not an artifact of layer 20. Table~\ref{tab:layerscan} scans SAE top800 across layers 10--30. Target effect grows smoothly with depth, but MMLU and GSM8K stay at baseline at every layer (baseline MMLU 0.692, GSM8K 0.700). A clean, capability-preserving medium-\(k\) handle exists throughout the mid-to-late stack, not at a single hand-picked layer.

\begin{table}[htbp]
\centering
\caption{Layer scan of SAE top800 on Gemma-2-9B (three seeds; layer 20 row uses six seeds). Target effect rises with depth while capability stays at baseline (MMLU 0.692, GSM8K 0.700) at every layer.}
\label{tab:layerscan}
\begin{tabular}{lcccc}
\toprule
Layer & Perturbation & True jailbreak & MMLU & GSM8K \\
\midrule
L10 & 0.406 & 0.075 & 0.681 & 0.733 \\
L15 & 0.311 & 0.175 & 0.693 & 0.713 \\
L20 & 0.205 & 0.170 & 0.683 & 0.715 \\
L25 & 0.277 & 0.275 & 0.700 & 0.720 \\
L30 & 0.234 & 0.290 & 0.680 & 0.737 \\
\bottomrule
\end{tabular}
\end{table}

\paragraph{The clean regime transfers to another architecture.} Table~\ref{tab:matched-pn-llama} repeats the matched-PN comparison on Llama-3.1-8B-Instruct with a Llama Scope SAE, now over six seeds. The qualitative picture matches Gemma-2-9B and is in fact stronger: at matched perturbation, SAE produces far more coherent harmful compliance than dense steering ($+0.30$ to $+0.38$, all significant), while MMLU and GSM8K are indistinguishable from zero except for a small GSM8K cost at the highest bin. The advantage is not an artifact of an under-powered dense baseline: extending the dense sweep to $\beta=0.60$---a per-token perturbation of $0.77$, more than double the SAE settings---still yields a true-jailbreak rate of only $0.057$, so dense never reaches the SAE jailbreak regime at any perturbation we tried. The advantage over all-layer dense steering is therefore not a Gemma-specific or Gemma-Scope-specific phenomenon. This Llama comparison, like the matched-PN result on Gemma-2-9B (Table~\ref{tab:matched-pn-9b}), is measured against an all-layer dense baseline; we did not re-run the surface-matched baselines of Table~\ref{tab:matched-surface-9b} at this scale.

\begin{table}[htbp]
\centering
\caption{Matched perturbation-norm comparisons on Llama-3.1-8B-Instruct (Llama Scope), SAE minus dense, paired bootstrap over six seeds. Units: true jailbreak 960, MMLU 3000, GSM8K 600. $^{*}$CI excludes zero.}
\label{tab:matched-pn-llama}
\resizebox{\linewidth}{!}{%
\begin{tabular}{llccc}
\toprule
Perturbation bin & Pair (SAE\,/\,dense) & True jailbreak & MMLU & GSM8K \\
\midrule
\(\approx 0.15\) & top800 / \(\beta\,0.05\)   & $+0.303^{*}$ [0.274, 0.333] & $-0.003$ [-0.015, 0.010] & $+0.020$ [-0.015, 0.055] \\
\(\approx 0.20\) & top1600 / \(\beta\,0.075\) & $+0.348^{*}$ [0.318, 0.379] & $-0.008$ [-0.020, 0.005] & $+0.033$ [0.000, 0.067] \\
\(\approx 0.30\) & top3200 / \(\beta\,0.15\)  & $+0.377^{*}$ [0.347, 0.408] & $-0.009$ [-0.023, 0.005] & $-0.062^{*}$ [-0.098, -0.025] \\
\bottomrule
\end{tabular}%
}
\end{table}

\paragraph{The advantage strengthens at 27B.} Scaling up to Gemma-2-27B-it sharpens the large-model picture on the safety, coherence, and perturbation axes. Table~\ref{tab:scale-27b} reports the matched coherence-gated results on the safety, coherence, and perturbation axes. Here SAE no longer merely wins at matched perturbation---it \emph{Pareto-dominates} dense steering. SAE top1600 reaches a true-jailbreak rate of $0.75$ at coherence $0.98$ with a relative perturbation of only $0.28$, whereas the smallest dense perturbation in our grid ($\beta=0.05$) is already more than twice as large ($0.60$) yet reaches a \emph{lower} jailbreak rate ($0.43$) at \emph{lower} coherence ($0.88$); pushing dense harder only collapses coherence further, to $0.16$ at $\beta=0.075$ and $0.06$ at $\beta=0.15$. The random-SAE control stays at baseline. Thus the perturbation-efficiency and coherence-retention advantage does not merely persist but \emph{grows} from 9B to 27B. We do not report 27B capability deltas: this checkpoint is loaded in 4-bit and its baseline GSM8K is already at floor ($0.12$) with MMLU degraded ($0.59$), so the capability axis is not comparable to the full-precision 9B/2B runs and we restrict the 27B claim to safety, coherence, and perturbation. As with the 9B and Llama matched-PN comparisons, the 27B dense baseline is all-layer; we report this Pareto-dominance as a trend relative to all-layer dense steering.

\begin{table}[htbp]
\centering
\caption{Matched coherence-gated results on Gemma-2-27B-it (4-bit), three seeds. SAE Pareto-dominates dense: it reaches a higher coherent-jailbreak rate at higher coherence and lower perturbation. Dense cannot enter the same regime---its smallest perturbation already exceeds SAE's while collapsing coherence as it is pushed. Capability is omitted because the 4-bit baseline is at floor (see text).}
\label{tab:scale-27b}
\begin{tabular}{lccc}
\toprule
Method & Perturbation (rel-norm) & True jailbreak & Coherence \\
\midrule
Baseline            & 0.000 & 0.006 & 1.000 \\
SAE top400          & 0.301 & 0.269 & 0.960 \\
SAE top800          & 0.289 & 0.533 & 0.975 \\
SAE top1600         & 0.275 & 0.752 & 0.977 \\
Dense \(\beta\,0.05\)   & 0.603 & 0.433 & 0.879 \\
Dense \(\beta\,0.075\)  & 0.845 & 0.073 & 0.163 \\
Dense \(\beta\,0.15\)   & 1.479 & 0.050 & 0.056 \\
Random SAE top1600  & 0.020 & 0.023 & 1.000 \\
\bottomrule
\end{tabular}
\end{table}

\paragraph{The clean regime breaks at small scale.} The same recipe fails on Gemma-2-2B-it. Table~\ref{tab:matched-pn-2b} shows that at matched perturbation the 2B SAE still raises the safety judge's jailbreak rate above dense, but at a severe capability cost that is absent in the large models: MMLU drops by 0.12--0.19 and GSM8K by 0.44--0.50, all significant, at every bin. A handle that destroys roughly half of GSM8K accuracy is not a localized control mechanism. This is a sharp negative control: the same feature-ablation procedure that is clean in 8B/9B models is destructive in a 2B model, so SAE-steering results obtained on small models should not be extrapolated to larger ones.

\begin{table}[htbp]
\centering
\caption{Matched perturbation-norm comparisons on Gemma-2-2B-it, SAE minus dense, paired bootstrap over six seeds. The jailbreak ``advantage'' coincides with a large, significant collapse of MMLU and GSM8K. Units: true jailbreak 960, MMLU 3000, GSM8K 600. $^{*}$CI excludes zero.}
\label{tab:matched-pn-2b}
\resizebox{\linewidth}{!}{%
\begin{tabular}{llccc}
\toprule
Perturbation bin & Pair (SAE\,/\,dense) & True jailbreak & MMLU & GSM8K \\
\midrule
\(\approx 0.15\) & top400 / \(\beta\,0.05\)  & $+0.174^{*}$ [0.144, 0.205] & $-0.123^{*}$ [-0.141, -0.106] & $-0.443^{*}$ [-0.483, -0.403] \\
\(\approx 0.20\) & top800 / \(\beta\,0.075\) & $+0.120^{*}$ [0.087, 0.153] & $-0.155^{*}$ [-0.175, -0.135] & $-0.502^{*}$ [-0.542, -0.460] \\
\(\approx 0.30\) & top1600 / \(\beta\,0.15\) & $+0.060^{*}$ [0.019, 0.102] & $-0.188^{*}$ [-0.209, -0.167] & $-0.502^{*}$ [-0.543, -0.462] \\
\bottomrule
\end{tabular}%
}
\end{table}

\begin{figure}[htbp]
\centering
\includegraphics[width=\linewidth]{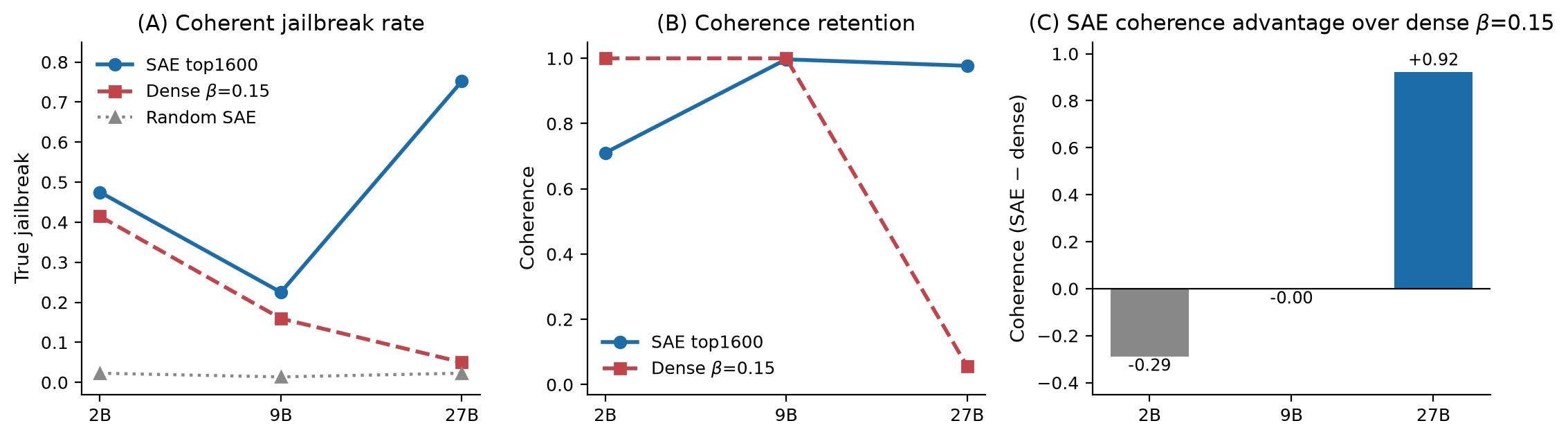}
\caption{Matched perturbation-norm summary across model scale (Gemma-2 2B/9B/27B). (A) True-jailbreak rate: SAE top1600 versus dense and the random-SAE control. (B) Coherence retention: SAE holds coherence while dense collapses as it is pushed, increasingly so at 27B. (C) SAE coherence advantage over dense $\beta=0.15$: near zero at 2B and 9B, large and positive at 27B. The clean regime strengthens upward from 9B to 27B; the small-scale capability failure at 2B is shown separately in Table~\ref{tab:matched-pn-2b}.}
\label{fig:scale-curve}
\end{figure}

\begin{figure}[htbp]
\centering
\includegraphics[width=\linewidth]{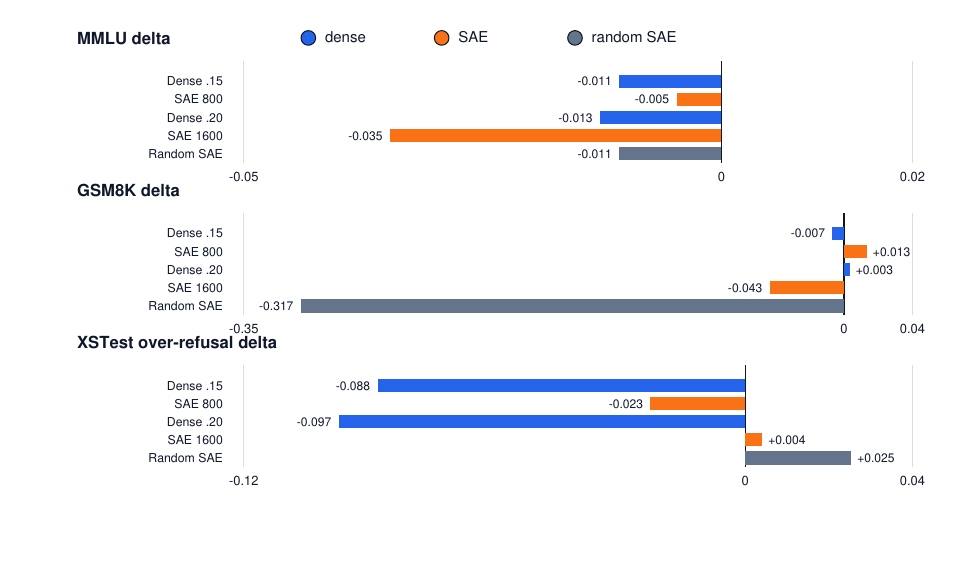}
\caption{Capability sensitivity on Gemma-2-9B. Random SAE top1600 strongly damages GSM8K, showing that large sparse-feature ablations can disrupt reasoning even when headline safety metrics look benign.}
\label{fig:capability}
\end{figure}

\paragraph{Sparse is not automatically safe.} Figure~\ref{fig:capability} shows that random SAE top1600 ablation severely damages GSM8K on the 9B model, even when MMLU does not fully expose the damage. This negative control reinforces the same message at fixed scale: large sparse-feature ablations are not inherently safe merely because they are sparse. The clean regime depends on selecting a feature set aligned with the target behavior, in a model large enough for that head to be separable from the computation that supports reasoning. A held-out benign retain set (Table~\ref{tab:l4-retain}) replicates this and adds a side-effect check: SAE top800 holds capability and benign over-refusal at baseline (MMLU 0.687, GSM8K 0.713, over-refusal 0.005 vs.\ baseline 0.007), SAE top1600 incurs only a small capability cost, and random SAE top1600 again collapses GSM8K (to 0.38)---yet no intervention raises benign over-refusal on the retain set, so the damage is concentrated in reasoning rather than in indiscriminate refusal.

\begin{table}[htbp]
\centering
\caption{Held-out benign retain set on Gemma-2-9B (three seeds, retain $n=141$). The aligned medium-$k$ head (SAE top800) preserves capability and does not increase benign over-refusal; the random-SAE control collapses GSM8K, replicating the capability story of RQ3 on retain data.}
\label{tab:l4-retain}
\begin{tabular}{lccc}
\toprule
Method & Benign over-refusal & MMLU & GSM8K \\
\midrule
Baseline            & 0.007 & 0.692 & 0.700 \\
Dense \(\beta\,0.075\)  & 0.007 & 0.683 & 0.727 \\
SAE top800          & 0.005 & 0.687 & 0.713 \\
SAE top1600         & 0.009 & 0.657 & 0.657 \\
Random SAE top1600  & 0.014 & 0.681 & 0.383 \\
\bottomrule
\end{tabular}
\end{table}

\FloatBarrier

\subsection{RQ4: Coherence Gating Is Necessary, and a Second Judge Confirms Large-Model Jailbreaks While Exposing Small-Model Inflation}

\paragraph{Human audit of the coherence gate.} We manually audit 101 sampled outputs from key categories to test whether the coherence gate separates meaningful harmful compliance from degenerate artifacts. This is a targeted, single-annotator audit of the most important gating failure mode rather than a blinded multi-annotator annotation study; we treat it as a sanity check on the gate, not as an exhaustive validation. The sample contains 30 SAE top3200 unsafe-incoherent outputs, 30 SAE top1600 unsafe-coherent outputs, 30 dense unsafe-coherent outputs, and 11 random-SAE unsafe or incoherent outputs. The audit supports the metric design. All 30 SAE top3200 unsafe-incoherent examples are judged degenerate rather than meaningful harmful compliance, while 25 of 30 SAE top1600 unsafe-coherent examples are judged harmful compliance. Figure~\ref{fig:audit} shows both the agreement of the coherence heuristic with human labels and the harmfulness breakdown within audit categories.

\begin{center}
\centering
\includegraphics[width=.86\linewidth]{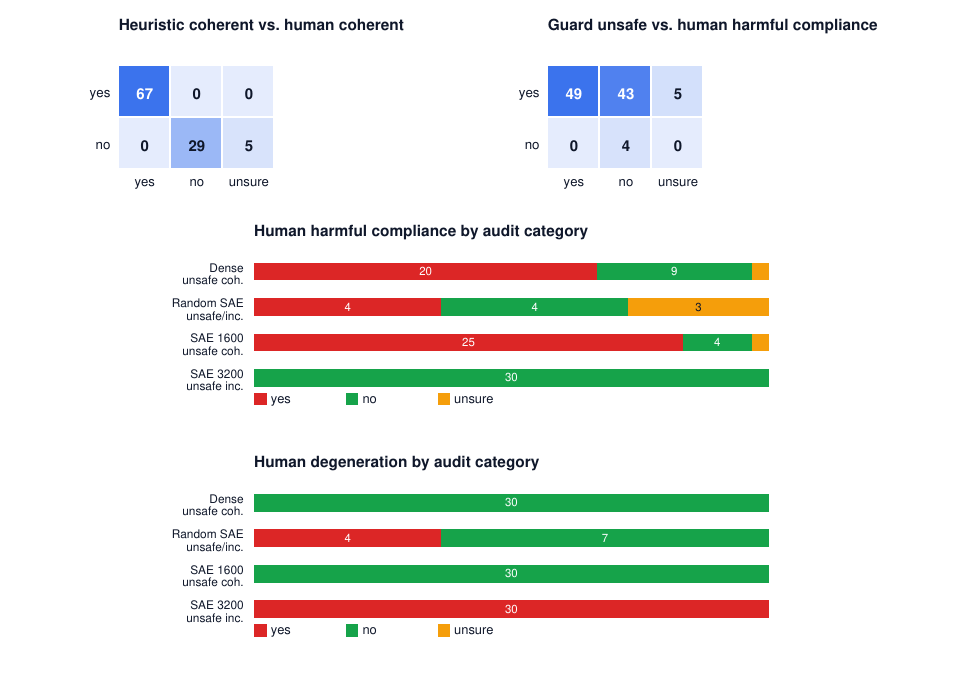}
\captionof{figure}{Targeted human audit of coherence and harmfulness labels (101 sampled Gemma-2-9B outputs, single annotator). The audit supports the coherence gate: high-strength SAE unsafe-incoherent examples are degenerate, while unsafe-coherent SAE examples are mostly meaningful harmful compliance.}
\label{fig:audit}
\end{center}

\paragraph{A second judge cross-checks the metric.} A single safety judge could still systematically mis-score harmful compliance, so we recompute the gated target metric with a HarmBench behavior-completion classifier and measure its agreement with Llama-Guard. Table~\ref{tab:second-judge} reports the two judges' gated jailbreak rates and their Cohen's \(\kappa\). On Gemma-2-9B the judges agree closely: the second judge's gated rates track Llama-Guard's at every point (e.g., SAE top400 0.147 vs.\ 0.138, \(\kappa=0.65\)), and both judges still place SAE above dense. The large-model true-jailbreak signal is therefore not an artifact of a single classifier.

\begin{table}[htbp]
\centering
\caption{Second-judge cross-check. ``Guard'' and ``HB'' are gated jailbreak rates under Llama-Guard and the HarmBench classifier; \(\kappa\) is their agreement. On 9B the judges agree (SAE confirmed). On 2B they diverge \emph{only} on SAE points, while agreeing on dense points; on Llama-3.1-8B---the same prompt source as 2B---the judges agree on SAE points, showing the 2B divergence is a real signal, not a template artifact.}
\label{tab:second-judge}
\begin{tabular}{llccc}
\toprule
Model & Method & Guard & HB & \(\kappa\) \\
\midrule
\multirow{4}{*}{Gemma-2-9B}
 & dense \(\beta\,0.05\)   & 0.038 & 0.025 & 0.59 \\
 & SAE top400              & 0.147 & 0.138 & 0.65 \\
 & SAE top800              & 0.172 & 0.128 & 0.58 \\
 & SAE top1600             & 0.222 & 0.166 & 0.38 \\
\midrule
\multirow{4}{*}{Gemma-2-2B}
 & dense \(\beta\,0.15\)   & 0.423 & 0.363 & 0.46 \\
 & SAE top400              & 0.302 & 0.121 & 0.26 \\
 & SAE top800              & 0.354 & 0.063 & 0.10 \\
 & SAE top1600             & 0.477 & 0.008 & 0.01 \\
\midrule
\multirow{3}{*}{Llama-3.1-8B}
 & SAE top800              & 0.310 & 0.291 & 0.43 \\
 & SAE top1600             & 0.355 & 0.347 & 0.38 \\
 & SAE top3200             & 0.390 & 0.368 & 0.35 \\
\bottomrule
\end{tabular}
\end{table}

\paragraph{Single-judge inflation on small-model SAE points.} On Gemma-2-2B the two judges diverge dramatically, but only on SAE points. For SAE top1600, Llama-Guard reports a gated jailbreak rate of 0.477 while the behavior-completion judge reports 0.008 (\(\kappa=0.01\)); for the matched 2B dense point the judges agree (0.423 vs.\ 0.363, \(\kappa=0.46\)). The small-model SAE ``jailbreaks'' are thus largely outputs that trip a surface safety judge without completing the harmful request---a second, independent line of evidence (beyond the capability collapse of RQ3) that the 2B SAE recipe is not producing genuine localized control. A natural concern is that the 2B and Llama runs use an in-the-wild jailbreak prompt source on which the HarmBench template is slightly off-specification. Llama-3.1-8B controls for this: it uses the \emph{same} prompt source, yet there the two judges agree on SAE points (\(\kappa\approx0.35\)--0.43 over six seeds, gated rates within 0.02). The divergence is therefore specific to small-model SAE outputs, not an artifact of the prompt source or judge template.

\subsection{RQ5: A Stable Refusal-Aligned Head Explains the Regime}

To understand why SAE top800 is more useful than larger top-\(k\) ablations, we reconstruct the layer-20 feature ranking for each split. The top-ranked features are refusal-aligned: they activate more on refused prompts than on harmful-compliance prompts. The separation decays rapidly with rank. Mean Cohen's \(d\) for complied-minus-refused activation is -0.5259 for ranks 1--50, -0.1256 for ranks 1--800, and only -0.0136 for ranks 2401--3200. Figure~\ref{fig:feature-diagnostics} summarizes this decay and the stability of the top800 feature set across splits.

No single feature is itself a jailbreak switch. Ablating individual top-ranked features one at a time, the most refusal-aligned feature (Cohen's \(d=-1.83\)) raises true jailbreak to at most $0.04$, and the mean single-feature effect over the top-ranked set is below $0.002$, with all outputs coherent. The useful control is therefore genuinely distributed over the medium-\(k\) head rather than carried by one dominant feature---consistent both with the rapid \(d\) decay and with why a single-feature or very-low-\(k\) intervention is too weak while a very-high-\(k\) one becomes fragile.

\begin{center}
\centering
\includegraphics[width=.88\linewidth]{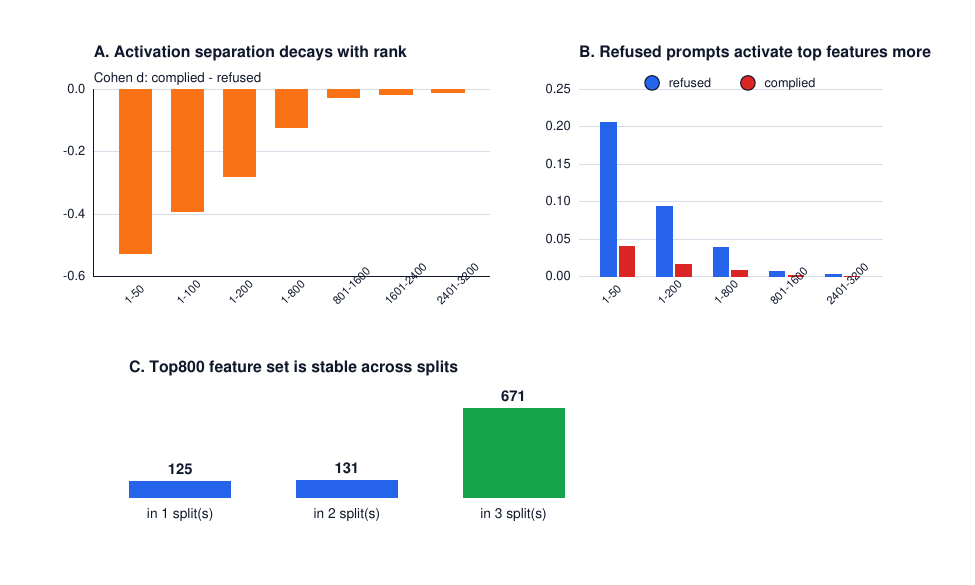}
\captionof{figure}{Feature-level diagnostics for the 9B layer-20 SAE. Top-ranked features are stable and refusal-aligned, while lower-ranked additions have weak activation separation. This explains why medium-\(k\) intervention is useful and high-\(k\) intervention becomes fragile.}
\label{fig:feature-diagnostics}
\end{center}

The top800 feature set is also stable across splits: 671 features appear in all three top800 sets, 131 appear in exactly two, and 125 appear in only one split. This supports the mechanism story that the useful regime is driven by a stable head of refusal-aligned features, while the tail adds weakly separated or incidental features that increase side-effect risk. The layer scan in Table~\ref{tab:layerscan} is consistent with this account: a refusal-aligned medium-\(k\) head is available throughout the mid-to-late stack, so the clean regime is a property of the ranking geometry rather than of one hand-picked layer. The feature diagnostics therefore connect the empirical regime boundary to the geometry of the SAE ranking rather than treating top-\(k\) as a purely empirical hyperparameter.

\subsection{RQ6: The Efficiency Advantage Is Direction-Dependent---Dense Wins for Refusal Injection}

The experiments so far intervene in the \emph{removal} direction: ablating refusal-aligned features to elicit harmful compliance. A deployment-facing safety intervention runs in the opposite direction---\emph{injecting} refusal to suppress harmful compliance---and it is not obvious that the SAE efficiency advantage transfers. We therefore repeat the matched-perturbation comparison on Gemma-2-9B in the safety-improving direction, comparing dense refusal addition (\texttt{dense\_add}, adding the refusal direction with coefficient \(\gamma\)) against SAE refusal-feature amplification (\texttt{sae\_amp}, boosting the top-\(k\) refusal features). We interpolate each method's harmful-compliance and capability as a function of per-token relative perturbation within the coherent region and compare the two on a common perturbation grid (three seeds); Table~\ref{tab:l2-pairing} reports the result.

The picture reverses relative to the removal direction. At matched perturbation, SAE amplification does suppress harmful compliance more aggressively than dense addition---but it does so by destroying reasoning: GSM8K falls from a baseline of $0.70$ toward $0.04$ at the strongest matched bin, while dense addition holds GSM8K essentially flat ($\approx 0.68$) across the entire range, with MMLU showing the same contrast. Read off at a fixed safety target of harmful compliance \(\le 0.30\), dense reaches the target at relative perturbation \(\approx 0.30\) for a negligible capability cost (GSM8K $-0.03$, MMLU $-0.01$), whereas SAE reaches it at \(\approx 0.24\) but at a heavy cost (GSM8K $-0.19$, MMLU $-0.06$). Both methods keep coherence near $1.0$ in this region, so the difference is a genuine capability cost, not a degeneration artifact. For the injection direction, dense steering is the perturbation-efficient choice and SAE amplification over-suppresses.

\begin{table}[htbp]
\centering
\caption{Matched perturbation-norm pairing in the \emph{safety-improving} (refusal-injection) direction on Gemma-2-9B, dense\_add vs.\ sae\_amp, interpolated within the coherent region (three seeds; baseline harmful compliance $0.908$, MMLU $0.692$, GSM8K $0.700$). At matched perturbation SAE suppresses harm more but collapses reasoning, whereas dense holds capability flat. Lower harmful compliance is safer; higher MMLU/GSM8K is better.}
\label{tab:l2-pairing}
\begin{tabular}{lcccccc}
\toprule
 & \multicolumn{2}{c}{Harmful compliance} & \multicolumn{2}{c}{GSM8K} & \multicolumn{2}{c}{MMLU} \\
\cmidrule(lr){2-3}\cmidrule(lr){4-5}\cmidrule(lr){6-7}
Perturbation (rel-norm) & dense & sae & dense & sae & dense & sae \\
\midrule
0.24 & 0.363 & 0.173 & 0.671 & 0.513 & 0.683 & 0.633 \\
0.30 & 0.291 & 0.117 & 0.675 & 0.398 & 0.682 & 0.582 \\
0.36 & 0.227 & 0.085 & 0.677 & 0.266 & 0.680 & 0.533 \\
0.42 & 0.181 & 0.052 & 0.677 & 0.135 & 0.678 & 0.485 \\
0.47 & 0.142 & 0.004 & 0.677 & 0.041 & 0.677 & 0.424 \\
\bottomrule
\end{tabular}
\end{table}

\begin{figure}[htbp]
\centering
\includegraphics[width=\linewidth]{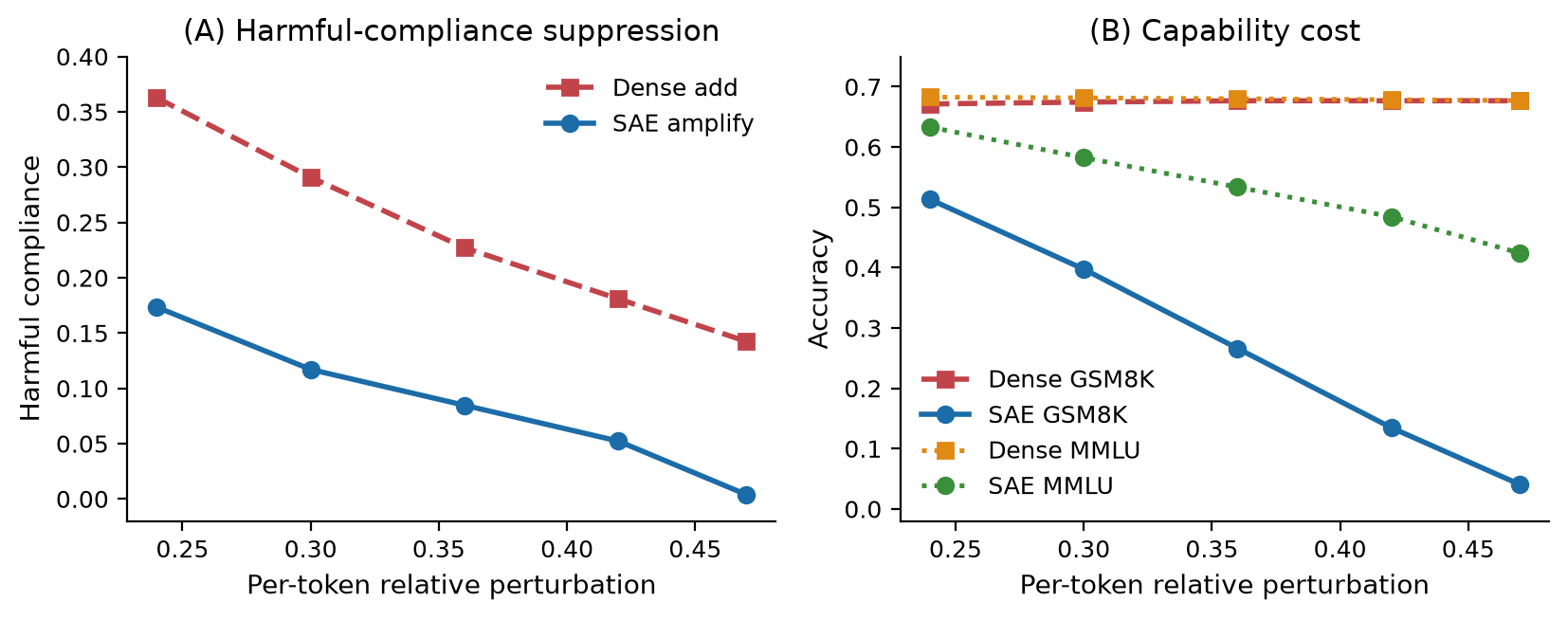}
\caption{Direction-dependent matched-perturbation pairing on Gemma-2-9B in the safety-improving (refusal-injection) direction. (A) Harmful-compliance suppression: at matched perturbation SAE amplification suppresses harm more aggressively than dense addition. (B) Capability cost: dense addition holds GSM8K and MMLU flat across the range, while SAE amplification collapses reasoning. The safety gain of SAE injection is therefore bought with a large capability cost that dense avoids---the reverse of the removal-direction result.}
\label{fig:l2-pairing}
\end{figure}

Taken together with RQ2--RQ3, this yields a direction-dependent account: sparse feature ablation is the more surgical handle for \emph{removing} a behavior, while dense steering is the more capability-preserving handle for broadly \emph{injecting} one. The right tool depends on the sign of the intervention, not on sparsity alone.

\section{Discussion}

Our results suggest that whether SAE interventions look localized is decided largely by the dense baseline they are matched against. Under a naive match of total perturbation norm, a medium-\(k\) SAE ablation on Gemma-2-9B reaches the same coherent harmful-compliance rate as all-layer dense steering with a smaller total perturbation---the signature usually read as localized control. But this signature does not survive matching the intervention \emph{surface}: against a same-layer dense baseline, and against a dense baseline restricted to the SAE decoder span, the advantage disappears and reverses (Table~\ref{tab:matched-surface-9b}), because the all-layer baseline was spending the same total perturbation thinly across the stack. On Llama-3.1-8B and Gemma-2-27B the advantage over all-layer dense is much larger and, at 27B, accompanied by a coherence-retention edge; we report these against an all-layer dense baseline. The comparison is also tied to the \emph{sign} of the intervention: for refusal injection (RQ6) dense steering is the more capability-preserving handle. Taken together, sparsity per se is not what makes an intervention clean---surface, basis, and direction matter more, and an evaluation that fixes only the total perturbation norm can read an all-layer-versus-single-layer mismatch as a localization advantage.

The small-model failure is an independent caution, on a different axis from the baseline-matching result. The same procedure that appears clean against an all-layer dense baseline in 8B/9B models is destructive in a 2B model: it halves GSM8K accuracy and its apparent jailbreak advantage is largely a single-judge artifact. Two independent diagnostics---capability collapse and judge disagreement---point the same way, and a same-source cross-architecture control rules out the obvious confounds. This is a cautionary result for a common practice in the interpretability literature: validating steering or feature-ablation methods on small models and assuming the conclusions transfer. They may not, and the failure can be hidden if evaluation relies on a single safety judge without a coherence gate and a capability check.

These findings also clarify how dense steering should be used as a baseline. Dense refusal directions are not straw methods; prior work shows that refusal can be strongly represented in activation space \citep{arditi2024refusal}. A fair SAE evaluation should therefore bracket the comparison from both sides---fixing behavior and fixing perturbation---because the two controls answer different questions; and crucially the dense baseline must share the SAE's intervention \emph{surface} and \emph{basis}, not merely its total perturbation norm, or an all-layer-versus-single-layer mismatch will masquerade as an efficiency advantage. For practical safety evaluation, the main lesson is that unsafe-only metrics are too weak for intervention studies. Automated judges remain useful, but their labels should be paired with coherence checks, a second behavioral judge, and a capability probe, especially when interventions or model scale push generations off-distribution.

\section{Limitations}
\label{sec:limitations}

The perturbation metric is a proxy for locality, not a proof of causal isolation. A smaller residual change can still affect distributed computations, and a larger dense change can sometimes preserve the task-relevant circuit. Our claim is therefore about a measurable behavior-per-perturbation comparison and its baseline dependence, not about a verified mechanistic circuit.

We previously left two stronger baselines---dense steering projected onto the SAE decoder span, and same-layer (rather than all-layer) dense steering matched to the SAE intervention surface---to future work; we run both on Gemma-2-9B (Table~\ref{tab:matched-surface-9b}), where they overturn the headline advantage. The Llama-3.1-8B and Gemma-2-27B comparisons in Section~\ref{sec:results} are reported against an all-layer dense baseline, and we label them as such throughout.

The cross-architecture and cross-scale comparisons are not perfectly balanced. On Llama-3.1-8B the dense refusal direction barely induces jailbreak at any strength: even extending the sweep to \(\beta=0.60\) (per-token perturbation \(0.77\)) leaves dense at a true-jailbreak rate of \(0.057\), so the matched-PN comparison there reflects equal perturbation rather than equal target effect because no dense target-effect match exists. We read this as a genuine property of the Llama refusal geometry rather than an under-powered baseline, but it does mean the Llama comparison cannot be made fully symmetric with the Gemma matched-TE analysis. The Gemma-2-27B configuration uses three seeds and is loaded in 4-bit, with capability benchmarks at or near floor at baseline (GSM8K $0.12$, MMLU $0.59$), so the 27B evidence is restricted to the safety, coherence, and perturbation axes; a full-precision 27B run would be needed to extend the no-capability-cost claim to that scale. The second judge uses a HarmBench behavior-completion template that is slightly off-specification on the in-the-wild jailbreak source used for the 2B and Llama runs; we mitigate this with a same-source cross-architecture control, but a native HarmBench behavior subset would further strengthen the second-judge analysis. The human audit is targeted and single-annotator rather than exhaustive or blinded: it supports the most important coherence-gating failure mode but does not replace a larger blinded multi-annotator study with reported inter-annotator agreement.

Finally, the core analysis intervenes by removing refusal-related features and measuring harmful compliance, a controlled setting for evaluating localization rather than a deployable safety-improving method. RQ6 takes one step toward the deployment-facing direction by comparing dense and SAE \emph{refusal injection}, and finds dense steering to be the more capability-preserving choice there, but a deployment-oriented intervention would still need to reduce harmful compliance while preserving helpfulness on genuinely benign traffic, and would require a different risk analysis than the held-out evaluation used here.

\section{Conclusion}

Whether SAE feature interventions look localized depends on how the dense baseline is matched. Under a matched coherence-gated protocol, the apparent perturbation-efficiency of SAE ablation on Gemma-2-9B survives a naive match of total perturbation norm but reverses once the dense baseline is restricted to the same layer or to the SAE decoder span---both fair baselines reach more coherent harmful compliance at matched perturbation, under two judges. The advantage over all-layer dense steering remains large on Llama-3.1-8B and at 27B. Independently of the dense comparison, high-\(k\) SAE ablations inflate a single safety judge with incoherent text, and in a 2B model the SAE jailbreak signal is largely single-judge inflation accompanied by capability collapse. We therefore argue for matched-surface, matched-basis, coherence-gated, multi-judge evaluation of SAE-based safety interventions, and we caution that apparent sparse-localization advantages can be artifacts of baseline specification rather than evidence of localized control.

\section*{Code and Data Availability}

\begin{sloppypar}
We will release a public artifact containing: the evaluation-protocol implementation and the intervention and analysis scripts, including the same-layer and decoder-span-projected dense baselines and the HarmBench second-judge rescorer used for the surface-matched comparison; the run configurations and an exact environment specification (pinned dependency versions, including \texttt{sae\_lens} 6.44.2); the model and SAE checkpoint identifiers with their resolved revisions/hashes, the 4-bit quantization configuration used for the Gemma-2-27B checkpoint, and the code commit used for every run; per-example evaluation metrics (\texttt{true\_jailbreak}, \texttt{mmlu\_correct}, \texttt{gsm8k\_correct}, coherence and judge labels) keyed by hashed prompt IDs, together with the aggregated result tables and bootstrap outputs; and the plotting code that produces every figure. To avoid redistributing harmful generations, we release the per-example metrics and hashed prompt IDs rather than the raw unsafe completions. Inputs that cannot be redistributed under their licenses (the in-the-wild jailbreak corpus used for the 2B and Llama runs) are replaced with regeneration scripts and a documented substitution procedure so the pipeline is reproducible end to end. Until the public release, the same materials are available from the authors on request.
\end{sloppypar}

\bibliographystyle{plainnat}
\bibliography{references}

\end{document}